# Social Assistive Robotics for Autistic Children

Stefania Brighenti
Adult Autism Center, DSM Local
Health Unit ASL
Turin, Italy
stefania.brighenti@edu.unito.it

Federico Buratto
Degree in Computer Science
University of Turin
Turin, Italy
federico.buratto@edu.unito.it

Fernando Vito Falcone
Degree in Computer Science
University of Turin
Turin, Italy
fernando.falcone@edu.unito.it

Cristina Gena
Department of Computer Science
University of Turin
Turin, Italy
cristina.gena@unito.it

Claudio Mattutino
Department of Computer Science
University of Turin
Turin, Italy
claudio.mattutino@unito.it

Matteo Nazzario
Intesa SanPaolo
Innovation Center
Turin, Italy
matteo.nazzario@intesasanpaolo.com

## ABSTRACT

This paper introduces the project *Social Assistive Robotics for Autistic Children* aimed at using robotic therapy for autism. The goal of the project is testing autistic children's interactions with the social robot NAO. In particular the robot will support the operators (psychologists, educators, speech therapists etc.) in their work. The innovative aspect of the project is that the children-robot interaction will consider the children's emotions and specific features and the robot will adapt its behavior accordingly.

CCS CONCEPTS

• Human-centered computing → User models • Human-centered computing → Accessibility systems and tools

## KEYWORDS

Socially Assistive Robotics, Human-Robot Interaction, Emotion recognition, Robotics for autism, Adaptive Robot

## 1 INTRODUCTION AND RELATED WORK

Autism Spectrum Disorders (ASD) are a group of neurodevelopmental disorders primarily characterized by deficits in social communication and interaction and by the presence of restricted and repetitive patterns of behaviors, interests or activities [1], [15].

Since the beginning of the 2000s, many researches have been carried out in order to better understand the potential use of social robots as a tool to promote interaction and communication in ASD. These studies have highlighted the potential benefits of social robotics to increase social and communicative behaviors in ASD, particularly in children and adolescents [16] [10][11].

Robotic therapy for autism has been explored as one of the first application domains in the context of socially assistive robotics (SAR), which aims to develop robots that help people with special needs through social interaction [16].

Previous studies ([3], [6], [12], [18]) showed the effects of social robots in improving emotional recognition and reciprocity, joint attention and triadic interaction [5], visual contact and social gaze [2] in ASD children.

Ribu [14] assumes that ASD students are attracted to computer science because computers are logical and consistent. Unlike social interactions that are often difficult to manage, interacting with robots and even programming are tasks that produce a set of expected outputs. ASD students prefer this type of problem-solving, in which their ability to organize large amounts of data and build reliable structures is useful and produces predictable results.

The NAO robot[1] has already been tested in clinical contexts with autistic children [20], [19]. The Aldebaran company (that formerly created NAO) supplied a set of API to use at best NAO in the therapy with autistic children[2], however, this project has been recently dismissed.

In the following, we will describe the project *Social Assistive Robotics for Autistic Children* concerning the use of robotic therapy for autism. The goal of the project is testing autistic children's interactions with the robot NAO. The robot will support the operators (psychologists, educators, speech therapists etc.) in their work. The innovative aspect is that the interaction with the robot considers the children's emotions and features and the robot will adapt its behavior accordingly. *Recognizing emotions* refers to the ability to identify and recognize the different types of emotions and the ways they are expressed through (i.e., expressed by face or body or voice). This ability is essential for interpersonal relations and is an important element for empathy, communication and social skills. Several studies have shown that the ability to recognize emotion is compromised in individuals with ASD [9].

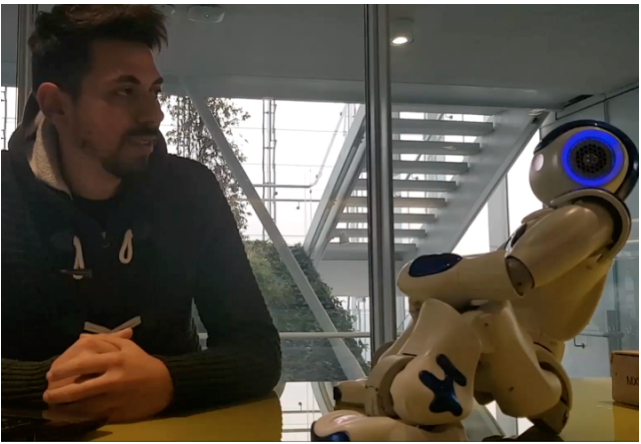

Figure 1: **NAO in an emotion detection session.**

## 2　THE MAIN GOAL OF THE PROJECT

The aim of this research project is to test the exchanges and interactions of ASD children with social robots. Specifically, the robot is intended to be an extra clinical tool for the operators (psychologists, educators, speech therapists, etc.). At the same time, robots may help operators in diagnosing and understanding autism. Indeed diagnosis may be improved through the use of both passive social cue measurement and interactions with a social robot to provide quantitative, objective measurements of social response [16]. One of the most innovative aspects of this research is the use of NAO improved with an emotion recognition software component aimed at recognizing the emotion in autistic children. This will allow us to recognize, through the robot, feedback from the children in addition to those from the clinical scales.

Moreover, we will enrich the intelligent behavior of NAO with a user modeling component [4], which will maintain a user model for every children considering his/her preferential choices [21]. The user model will be initialized manually with the help of the operators, and will be dynamically updated according to the perceived emotions and behavior of the child. Thus NAO will be able to express a more sociable and intelligent behavior: it will be able to recognize the user, remember the past interactions, remember the past perceived emotion in its user, and at runtime it will be able to adapt its behavior on the current user mood.

This last point is linked to the adaptive [4] and persuasive mechanisms [22] with which the NAO behavior will be extended. We will enrich the robot intelligence with reasoning skills and knowledge base, which will allow the robot to adapt to user specific needs and features and it will customize its interaction and behavior accordingly. Therefore, the robot will become adaptive with respect to the user, thus able to adapt its behavior according to the different and evolving user needs, but it will be also adaptable (see [4]) since we will also provide operators a simplified end-user programming environment that will allow them to manually chose and eventually adapt the current script (e.g. a social story) to be performed by the robot during the therapy session. The same tool will record and analyze every therapy session and will be designed tighter with the operators in order to be a further help for their work.

In the following we listed the main phases and steps of the project:

**Phase 1** - **Basic emotions recognized by the robot**
- Integration in the NAO robot with external software for the recognition of basic emotions
- User testing (Note: at this stage, minor subjects and/or those with autism problems will not be involved)
- Redesign and tuning based on test results

**Phase 2** - **Basic emotions recognized by the robot**
- Study and analysis of emotion recognition software in autistic children
- Integration in the NAO robot of external software for the recognition of the emotions of autistic children
- Experimental activities will be defined with the collaboration of the ethics committee of the University of Turin

**Phase 3** - Implementation of adaptive behavior in the robot. This phase will depend strongly from the results obtained in

---
[1] https://www.softbankrobotics.com/emea/en/nao
[2] http://doc.aldebaran.com/2-1/asknao/index_asknao.html

the course of Phase 2. Therefore, the adaptive behavior of the robot can be more or less sophisticated according to the trend of the results recorded in Phase 1 and 2
- Study and analysis of a user model for target subjects (autistic children) and its implementation
- Study and analysis of knowledge bases, of the reasoning and learning mechanisms of the robot
- Study, analysis and implementation of an end-user environment the will allow the operator to configure and manually adanpt the performance and behavior of the robot
- Study, analysis and implementation of script and story to be performed by the robot
- Integration of the main software components
- Definition of an experimental protocol
- Final testing

Our idea is to store all the main software components on the cloud, and use the robot as client sending request and receiving response for adapting its behavior.
In all the experimental activities listed above we will involve a specialized structure in the diagnosis and treatment of patients with autism spectrum disorders: Fondazione Paideia[3].

## 3   INITIAL STEPS

As a first step, in order to familiarize with the use of the emotion recognition software and their integration into NAO, we started working with external software that recognize the basic emotions in neurotypical subjects and then we integrated it in NAO.
We are now in the middle of Phase 1. We have integrated in the NAO robot two software components: one for the recognition of basic emotions by Facial Expression Recognition (FER) approaches, and another one performing sentiment analysis of the sentences spoken by the users to the robot.
For the first component, we decided to use two existing FERs: PrimEmo, a FER developed by the Computer Science Department of the University of Bari able to recognize the six basic emotions from Ekman (happiness, sadness, surprise, fear, anger, disgust [7]) and the face-detect API from Microsoft Azure platform[4]. In order to have to NAO robot contacting the two external services we used we used a Raspberry PI serving as middleware.
The final architecture of the resulting RES system consists of five phases:
1. The Raspberry PI sends the script that contacts the PrimEmo or Microsoft Azure service running and waits to receive requests;
2. Nao captures the image, sends it via socket to the software running on the Raspberry PI and waits for a response;
3. The software on Raspberry PI receives the image and sends it to the preset service for analysis;
4. The software on Raspberry PI receives a JSON with confidence intervals related to the emotions encountered, adds an extra field to the JSON specifying the type of service used for detecting the emotion (PrimEmo or Azure) and sends the JSON to the robot via socket. If the service has not been able to detect a face, a JSON is sent containing a message error;
5. The robot receives the message, calculates the maximum between the values of the emotions received, and based on the predominant emotion performs a different animation. If the received JSON contains the value message error, the robot will say that it failed to recognize the emotion of the user and it will ask the user will try to repeat the task (please notice that in this phase we are testing the emotion recognition with neurotypical user and demonstration tasks).

At the moment we are in contact with the University of Bari to try to improve the current version of PrimEmo software, so as to better adapt it to the needs of future users. Furthermore, in the future, numerous interaction scenarios could be developed between the user and NAO robot that exploit the recognition of emotions, both in the therapeutic and in the care fields.
 For the second component (sentiment analysis of the spoken sentences) we created a service for the robot being able to transform the sentences spoken by the user into text, analyze it, understand its sentiment and choose the most suitable answer at that moment.
 As for the previous component, we decided to use a Rasperry PI as a bridge between the robot and the external cognitive services platform. This allowed the decoupling of the robot specific characteristics from the peculiarity of the integrated AI services, avoiding also performance problem due to limited computational capacity of the robot. The Azure Cognitive Services[5] suite from Microsoft was chosen for the sentiment analysis and speech-to-text service. While Python was chosen as the programming language, being the language with greater compatibility with the robot along with C#.
The Microsoft Azure Cognitive Service uses machine learning based algorithm for sentiment analysis classification. From this process it outputs a value between 0 and 1: the higher the value, the more positive the sentiment will be, conversely, the more this value approaches zero and the more it will be negative. To perform this analysis, various techniques are used that are combined with each other, including: the association of words, the analysis of parts of the speech and the text processing.

---

[3] https://www.fondazionepaideia.it/

[4] https://azure.microsoft.com/en-us/services/cognitive-services/face/

[5] https://docs.microsoft.com/en-us/azure/cognitive-services/welcome

Concerning our implementation, as soon as the user begins to speak, the robot creates multiple recordings interspersed with speech pauses, which will be first transformed into text and then analyzed. In order to that, four Python scripts have been implemented:

- **rms.py**: executed by the robot, it connects to rec.py as client using a socket and calculate the RMS (Root Mean Square[6]) every 0.1 seconds to understand if a user is speaking with the robot. When this happens, it sends a message to rec.py which will act accordingly;
- **rec.py**: executed by the robot, it connects to out.py as client and accepts connections, thus performing also the role of a server, from rms.py. It waits for messages from rms.py, to whom it communicates to start or stop a recording or to finish the execution. Once registration is complete, the script saves, fragments and sends the registration to out.py;
- **out.py**: executed on a hardware external to the robot (Rasperry PI), it connects as client to the Python script running on NAO and waits for rec.py making a connection request. It takes care of receiving the recordings, reassembling them and requesting the speech-to-text service. When it receives the end message, it requests the sentiment analysis output and sends everything to the Python script running on NAO;
- **Python script running on NAO**: executed by the robot, it establishes a connection with out.py and it expects to receive a string containing the words spoken by the user and the output of the sentiment analysis. He will then respond with a sentence related to the calculated sentiment.

The advantage of having multiple communicating programs is the parallel execution. This makes the robot's response much faster than running a single program that should translate a single larger recording at the end of the dialogue.

## 4 CONCLUSION

The project has just started. The examples reported above witnesses our first implementation steps. In the following, we will integrated the 2 components in order to jointly perform the two related analyses and outputs. We are also organizing testing with neurotypical users in order to evaluate the performance of the described software components.

Then, we will move towards the integration of a specific FER software for ASD children. Following a review of the literature, we will integrate existing software, evaluate them, and we will also implement our specific and tailored FER solution in order to compare the results.

## ACKNOWLEDGMENTS

The research was conducted under a cooperative agreement between University of Turin and Intesa Sanpaolo Innovation Center.

.

## REFERENCES

[1] American Psychiatric Association. (2013). Diagnostic and Statistical Manual of Mental Disorders (5th ed.)
[2] Admoni, H., Scassellati, B.: Social eye gaze in human-robot interaction: A review. Journal of Human-Robot Interaction 6(1), 25–63 (2017)
[3] Aresti-Bartolome N., Garcia-Zapirain B. (2014). Technologies as Support Tools for Persons with Autistic Spectrum Disorder: A Systematic Review.
[4] Brusilovsky, Peter (1996). "Methods and Techniques of Adaptive Hypermedia". *User Modeling and User-Adapted Interaction*. **6** (2–3): 87–129. doi:10.1007/bf00143964.
[5] Chevalier, P., Martin, J.C., Isableu, B., Bazile, C., Iacob, D.O., Tapus, A.: Joint attention using human-robot interaction: Impact of sensory preferences of children with autism. In: Robot and Human Interactive Communication (RO-MAN), 2016 25th IEEE International Symposium on. pp. 849–854. IEEE (2016)
[6] Diehl J.J., Schmitt L.M., Villano M., Crowell C.R. (2012). The Clinical Use of Robots for Individuals with Autism Spectrum Disorders: A Critical Review from http://files.eric.ed.gov/fulltext/EJ967456.pdf
[7] Ekman P. e W. Friesen, Facial Action Coding System: A Technique for the Measurement of Facial Movement, Palo Alto, 1978.
[8] Huijnen C., Lexis M., Jansens R., De Witte L. (2017). How to Implement Robots in Interventions for Children with Autism? A Co-creation Study Involving People with Autism, Parents and Professionals
[9] Lima, Antonio Marcos Oliveira de, Medeiros, Maxson Ramon dos Anjos, Costa, Paula Dornhofer Paro, & Azoni, Cíntia Alves Salgado. (2019). Analysis of softwares for emotion recognition in children and teenagers with autism spectrum disorder. Revista CEFAC, 21(1), e12318. Epub February 11, 2019.https://dx.doi.org/10.1590/1982-02162019/21112318
[10] Palestra G., Bortone I. (2016). Perspective Ethical Issues about Experiences with Social Robots to help Children with Autism Spectrum Disorders. New Friends 2016, Conference Proceedings. Ethical, Legal and Societal Issues of Robots in Therapy and Education Workshop.
[11] Palestra G., Esposito F., De Carolis B. (2017). A Multimodal Interface for Robot-Children Interaction in Autism Treatment, *DCPD@CHItaly 2017: 158-162*
[12] Pennisi P.,Tonacci A., Tartarisco G., Billeci L., Ruta L., Gangemi S., Pioggia G. (2016). Autism and social robotics: A systematic review. Autism research, https://doi.org/10.1002/aur.1527
[13] Qidwai, U., Kashem, S.B.A. & Conor, O. Humanoid Robot as a Teacher's Assistant: Helping Children with Autism to Learn Social and Academic Skills. *J Intell Robot Syst* (2019) doi:10.1007/s10846-019-01075-1
[14] Ribu, K. (2010). Teaching Computer Science to Students with Asperger's Syndrome. Proceedings from NIK- 2010: *The Norwegian*

---

[6] The root mean square is a value obtained from the sound waves felt by the microphones placed on Nao's head, this value allows to understand when a user starts or finishes talking to the robot. The value expresses the mean-root square pressure which is the root square of the average of the square of the pressure of the sound signal in a given duration.

*Informatics Conference.* Bergen, Norway. Retrieved from http://www.nik.no/2010/10-Ribu.pdf

[15] Safran, J. (2002). A Practitioner's Guide to Resources on Asperger's Syndrome. Intervention in School & Clinic, 37 (5), 283-298

[16] Scassellati, Brian. (2005). How Social Robots Will Help Us to Diagnose, Treat, and Understand Autism. Robotics Research. 28. 552-563. 10.1007/978-3-540-48113-3_47.

[17] Scassellati B, Henny Admoni, and Maja Matric. 2012. Robots for use in autism research. Annual review of biomedical engineering 14 (2012), 275–294.

[18] Seon-wha Kim E. (2013). Robots for social skills therapy in autism: evidence and designs toward clinical utility.

[19] S. Shamsuddin *et al.*, "Initial response of autistic children in human-robot interaction therapy with humanoid robot NAO," *2012 IEEE 8th International Colloquium on Signal Processing and its Applications*, Melaka, 2012, pp. 188-193. doi: 10.1109/CSPA.2012.6194716

[20] Adriana Tapus, Andreea Peca, Amir Aly, Cristina Pop, Lavinia Jisa, et al.. Children with Autism Social Engagement in Interaction with Nao, an Imitative Robot - A Series of Single Case Experiments. Interaction Studies, 2012, 13 (3), pp.315-347. ffhal-01265990f

[21] Anthony Jameson, Silvia Gabrielli, Per Ola Kristensson, Katharina Reinecke, Federica Cena, Cristina Gena, and Fabiana Vernero. 2011. How can we support users' preferential choice? In CHI '11 Extended Abstracts on Human Factors in Computing Systems (<i>CHI EA '11</i>). Association for Computing Machinery, New York, NY, USA, 409–418. https://doi.org/10.1145/1979742.1979620

[22] Gena, C.; Grillo, P.; Lieto, A.; Mattutino, C.; Vernero, F. When Personalization Is Not an Option: An In-The-Wild Study on Persuasive News Recommendation. *Information* **2019**, *10*, 300. https://doi.org/10.3390/info10100300